\documentclass[letterpaper]{article} % DO NOT CHANGE THIS
\usepackage[preprint]{aaai2027}  % DO NOT CHANGE THIS
\usepackage[hyphens]{url}  % DO NOT CHANGE THIS
\usepackage{graphicx} % DO NOT CHANGE THIS
\usepackage{natbib}  % DO NOT CHANGE THIS AND DO NOT ADD ANY OPTIONS TO IT
\usepackage{caption} % DO NOT CHANGE THIS AND DO NOT ADD ANY OPTIONS TO IT
\usepackage{algorithm}
\usepackage{algorithmic}
\usepackage{booktabs}
\usepackage{amsmath}
\usepackage{amssymb}
\usepackage{subcaption}
\usepackage{enumitem}
\usepackage{booktabs}
\usepackage{multirow}
\usepackage{graphicx}
\usepackage{array}
\usepackage{float}
\title{Threat-guided Policy-aware Scene Perturbation for Safe Autonomous Driving with Online Reinforcement Learning}
\author{
Xincong Hu\textsuperscript{\rm 1}\thanks{This work was done during the internship at Yinwang Intelligent Technology Co., Ltd.},
Lei Ou\textsuperscript{\rm 1}\footnotemark[1],
Maosen Li\textsuperscript{\rm 2}, \\
Jingtao Zhang\textsuperscript{\rm 2},
Liguo Hou\textsuperscript{\rm 2},
Zongzhang Zhang\textsuperscript{\rm 1}\thanks{Corresponding author: zzzhang@nju.edu.cn}
}

\affiliations{
\textsuperscript{\rm 1}Nanjing University, Nanjing, Jiangsu, China\\
\textsuperscript{\rm 2}Yinwang Intelligent Technology Co., Ltd., China
}

\begin{document}

\maketitle

\begin{abstract}
Reinforcement learning (RL) has shown promising performance in autonomous driving, yet ensuring the safety of online RL policies remains challenging due to insufficient exposure to safety-critical driving scenes. The long-tailed nature of real-world traffic situations makes dangerous and rare interactions difficult to encounter through conventional sampling, limiting the ability of RL policies to learn robust safety behaviors. Existing methods improve training diversity by synthesizing challenging scenes or adversarial situations. However, these approaches typically optimize scene generation objectives separately from the evolving policy, without explicitly modeling how generated perturbations relate to the current policy's weaknesses and learning needs. In this paper, we propose Threat-guided Policy-aware Scene Perturbation (TPSP) for safe autonomous driving with online RL. TPSP introduces a policy-aware scene encoder to capture the interaction between policy behaviors and surrounding environments, enabling scene perturbation aligned with the current policy. Based on this representation, TPSP selectively perturbs critical objects rather than applying uniform modifications across the scene. Furthermore, we develop a threat-guided optimization strategy that evaluates perturbed scenes through threat-level differences between policy rollouts on original and perturbed scenes, guiding the generation of safety-critical scenes with higher training value. Comprehensive experiments demonstrate that TPSP improves safety learning efficiency, achieving strong safety performance on NAVSIM v2 with approximately 4 million kilometers of simulated driving data. Ablation studies verify that policy-aware targeted perturbations provide more informative safety-critical experiences than random or policy-unaware strategies, enabling safer driving under limited interaction budgets.
\end{abstract}

\section{Introduction}

Reinforcement learning (RL) has emerged as a promising approach for
autonomous driving by enabling policies to learn complex decision-making
strategies through environment interaction
~\cite{kiran2022deep,cusumano-towner2025robust,kazemkhani2025gpudrive}.
Unlike rule-based systems relying on handcrafted objectives
~\cite{paden2016survey} and supervised approaches requiring expert
demonstrations~\cite{codevilla2018end,bansal2019chauffeurnet}, RL can
adapt driving behaviors through large-scale experience collection.
However, achieving safe and robust driving remains challenging due to
the scarcity of safety-critical long-tail scenes, where rare events
such as collisions provide insufficient training signals for learning
reliable safety behaviors.

To improve the exposure of RL policies to challenging driving scenes,
prior studies explore scene augmentation and perturbation strategies to
enrich training experiences. Representative approaches, including
prioritized replay~\citep{schaul2016prioritized}, curriculum learning
~\citep{bengio2009curriculum}, and adversarial scene generation
~\citep{xu2022safebench}, focus on selecting informative samples or
constructing failure-inducing scenes. However, existing methods
typically characterize scene difficulty based on environment-level
objectives, such as collision probability, adversarial objectives, or
predefined safety violations
~\cite{zhou2020learning,wang2021advsim,xu2022safebench,chen2025frea}.
Consequently, the generated scenes are often independent of the
evolving policy: a specific event, such as a cut-in or sudden braking,
may be trivial for a mature policy but challenging for an immature one.
This limitation prevents existing difficulty-oriented strategies from
effectively targeting the current policy weaknesses during online RL
training.

% Upload pipeline.pdf to the same Overleaf directory as this .tex file.
\begin{figure*}[t]
    \centering
    \includegraphics[width=\textwidth]{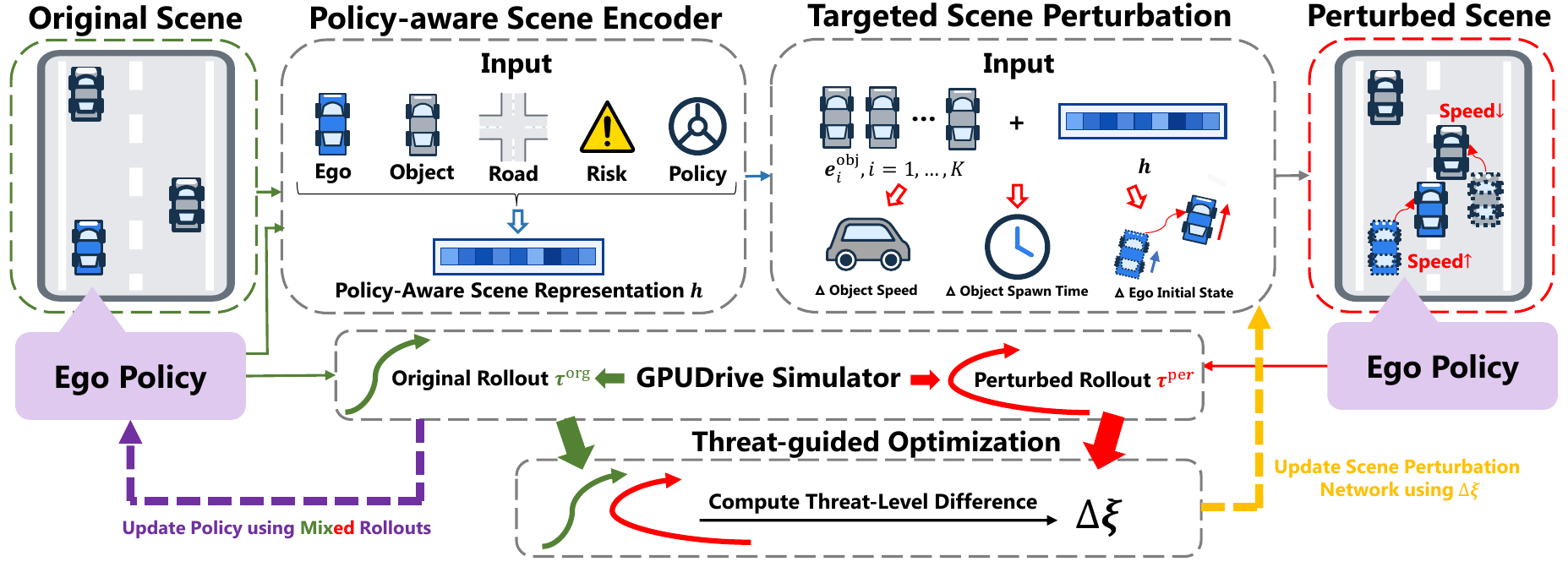}
    \caption{Overview of the proposed framework.}
    \label{fig:framework}
    \vspace{1mm}
\end{figure*}

Motivated by this observation, we propose that an effective scene perturbation should consider the current policy behavior instead of relying on predefined scene modifications. Since different policies may fail under different traffic conditions, policy-agnostic perturbations often provide limited training benefits. Therefore, a more effective perturbation strategy should adapt to current policy characteristics and identify the safety-critical situations that are most relevant to its limitations. Based on this insight, we propose TPSP, a Threat-guided Policy-aware Scene Perturbation framework for safe autonomous driving with online RL. TPSP identifies and perturbs policy-relevant challenging scenes, enabling more effective safety-oriented policy optimization.

As illustrated in Figure~\ref{fig:framework}, TPSP consists of three key components: a policy-aware scene encoder, a targeted scene perturbation module, and a threat-guided scene perturbation optimization method. First, the policy-aware scene encoder integrates ego states, surrounding objects, road context, scene risk information, and the current policy's action distribution to construct a policy-aware scene representation. This representation provides both scene-level and policy-level information for perturbation generation. With this representation, targeted scene perturbation selectively modifies critical objects instead of applying uniform perturbations across the scene. Finally, TPSP evaluates perturbed scenes by computing the threat-level differences between the policy rollouts on perturbed and original scenes. The resulting threat signal optimizes the scene perturbation network, guiding it toward discovering safety-critical scenes with higher training value for autonomous driving policy improvement.

Extensive experiments on autonomous driving benchmarks demonstrate that TPSP improves safety learning efficiency under limited online RL interaction budgets. With roughly 4 million kilometers of simulated driving data, TPSP effectively discovers policy-relevant safety-critical experiences and enhances policy robustness. On the NAVSIM v2 \texttt{navhard\_two\_stage} benchmark~\citep{cao2025navsimv2}, TPSP achieves 99.8\% NC and 99.6\% TTC on Stage 1 and maintains strong performance in the more challenging Stage 2 setting with 96.7\% NC and 94.6\% TTC, outperforming previous best results by 2.2\% and 1.8\%, respectively. These results demonstrate that policy-aware exploration enables more efficient utilization of limited interaction data for learning safer driving behaviors. In this work, our contributions are summarized as follows:
\begin{itemize}
    \item We propose TPSP, a novel threat-guided perturbation framework designed to generalize safety-critical scenes for safe autonomous driving. By targeting the current policy, TPSP generates high-risk interaction cases to robustly train safe driving behaviors.
    \item Within TPSP, we quantify the threat by comparing pre- and post-perturbation rollouts to identify high-risk samples. The scene perturbation module is then trained with the objective of maximizing these threat scores.
    \item Extensive experiments demonstrate that TPSP significantly outperforms baseline methods in both driving safety and data efficiency, highlighting its effectiveness in training robust driving policies.
\end{itemize}

\section{Related Work}

\paragraph{Reinforcement learning for autonomous driving.}
Reinforcement learning (RL) has been widely studied for closed-loop
autonomous driving, where policies are optimized through continuous
interaction with dynamic environments. To support RL-based training,
various simulation platforms have been developed to improve scene
diversity, scalability, and realism. CARLA~\citep{dosovitskiy2017carla}
and SMARTS~\citep{zhou2021smarts} provide high-fidelity simulation and
multi-agent interaction, while MetaDrive~\citep{li2022metadrive} and
GPUDrive~\citep{kazemkhani2025gpudrive} further enhance scene
generation and large-scale parallel experience collection. Recent works
also explore more realistic and effective RL training paradigms, such as
3DGS-based simulation in RAD~\citep{gao2025rad} and aligned world models
in Raw2Drive~\citep{DBLP:conf/nips/YangJLYYY25} for driving
policy optimization. However, RL policies remain highly dependent on the
quality and diversity of collected experiences, making rare and
safety-critical driving scenes difficult to discover through
conventional sampling.

\paragraph{Safety-critical scene generation and perturbation.}
Existing works improve the exposure of driving policies to challenging
situations through safety-critical scene generation. AdvSim~\citep{wang2021advsim} and KING
~\citep{hanselmann2022king} optimize surrounding-agent behaviors to
construct safety-critical interactions, while SafeBench
~\citep{xu2022safebench} and CAT~\citep{zhang2023cat} provide
benchmarking and critical scene generation frameworks for robustness
evaluation. Recent approaches further explore learning-based scene
construction, including feasibility-aware adversarial optimization
~\citep{chen2025frea}, diffusion-based safety-critical synthesis
~\citep{xu2025diffscene}, and collaborative adversarial scene
evolution~\citep{liu2026scenge}. However, these methods mainly optimize
scene difficulty, failure likelihood, or adversarial object behaviors
from environment-level objectives, without explicitly considering the
evolving policy and its specific learning needs. TPSP addresses this
limitation by introducing policy-aware representations for targeted
scene perturbation and optimizing perturbations with policy-relevant
threat signals.

\paragraph{Policy-aware reinforcement learning.}
Policy-aware lear\-ning improves training efficiency by adapting
experiences or interactions according to the current policy.
Self-play methods such as AlphaZero~\citep{silver2018alphazero}
demonstrate the effectiveness of informative interaction generation,
while several autonomous driving studies explore adaptive agent
interactions and robust policy optimization
~\citep{dai2023socially,cao2023continuous,cusumano-towner2025robust}.
Recent RL driving frameworks such as CaRL~\citep{jaeger2025carl} and
adaptive curriculum learning~\citep{abouelazm2025automatic} further
adjust training experiences based on policy capability. However, these
methods focus mainly on policy optimization or experience scheduling,
rather than constructing safety-critical scenes tailored to the
weaknesses of current policy. In contrast, TPSP establishes a
closed-loop policy-aware perturbation process, where the policy guides
scene perturbation to provide more valuable experiences for online RL
training.

\section{Problem Formulation}

We formulate closed-loop autonomous driving as a Markov decision process (MDP), $\mathcal{M}=(\mathcal{S},\mathcal{A},\mathcal{P},\mathcal{R},\gamma)$, where $\mathcal{S}$, $\mathcal{A}$, $\mathcal{P}$, $\mathcal{R}$, and $\gamma$ denote the state space, action space, transition function, reward function, and discount factor, respectively. At timestep $t$, the driving policy $\pi_{\theta}(a_t|s_t)$ maps the current state $s_t\in\mathcal{S}$ to an action $a_t\in\mathcal{A}$. The state contains observable information of the ego vehicle, surrounding objects, and driving environment.

A driving policy rollout is represented as
\begin{align}
\tau=(s_0,a_0,r_0,\ldots,s_T),
\end{align}
where $r_t\in\mathcal{R}$  is the reward received at timestep $t$ and $T$ represents the rollout horizon.

To generate policy-relevant safety-critical scenes, we introduce a scene perturbation network $G_{\phi}$ parameterized by $\phi$. Given a policy-aware scene representation, $G_{\phi}$ predicts perturbations including object speed adjustment, object spawn time shift, and ego initial state modification.

The objective of TPSP is to optimize $G_{\phi}$ to discover safety-critical scenes with higher training value for policy optimization. Specifically, TPSP updates $G_{\phi}$ using the threat-level difference between policy rollouts on perturbed and original scenes as the optimization signal.

\section{Method}
The proposed TPSP framework consists of a driving policy $\pi_{\theta}$
and a scene perturbation network $G_{\phi}$. Given policy-aware scene
representations, $G_{\phi}$ generates targeted scene perturbations, which
are used to collect rollouts for policy optimization. The perturbations
are optimized through policy-relevant threat signals derived from the
comparison between policy rollouts on perturbed and original scenes, forming a closed-loop process that enables efficient learning of robust safety behaviors.

\subsection{Policy-aware Scene Encoder}

The effectiveness of scene perturbation depends not only on the traffic situations, but also on how the current policy interacts with the scene. Therefore, TPSP learns a policy-aware scene representation that integrates traffic configuration information and policy characteristics, enabling the scene perturbation network to generate perturbations aligned with the current policy. Given a scene state $s_t$, TPSP extracts heterogeneous features from the scene, including ego feature, surrounding object features, road feature, risk-related feature and policy feature conditioned on $s_t$.

The ego feature $\boldsymbol{e}^{\mathrm{ego}}$ encodes the current kinematic state of the ego vehicle, including velocity, heading, and destination information. For the $i$-th surrounding object, we define its feature representation as $\boldsymbol{e}^{\mathrm{obj}}_i, i=1,\dots,N$, where $N$ denotes the number of observable objects. Each object feature contains intrinsic attributes, such as object type and geometric properties, as well as interaction-related information with the ego vehicle, including relative position, relative velocity, and Time-to-Collision (TTC). These interaction features characterize the dynamic relationship between the ego vehicle and surrounding objects.

Moreover, the road feature $\boldsymbol{e}^{\mathrm{road}}$ encodes the information of the surrounding road structure, providing static environmental context for scene representation. In addition, TPSP incorporates a scene-level risk-related feature $\boldsymbol{e}^{\mathrm{risk}}$ to provide complementary safety information. This feature aggregates several safety-related indicators, including the inverse minimum distance to surrounding objects, the inverse minimum TTC, and the number of surrounding objects in the scene. 

To capture the characteristics of the current driving policy, TPSP extracts a policy feature from the hidden representation of the frozen policy network. Specifically, let $\pi_{\bar{\theta}}$ denote a frozen copy of the current driving policy during rollout collection, where $\bar{\theta}$ represents the fixed policy parameters. Given the original scene observation before perturbation, the hidden representation extracted from $\pi_{\bar{\theta}}$ is denoted as $\mathbf{z}_{\pi_{\bar{\theta}}}$. The policy feature is obtained through a projection network:
\begin{equation}
\boldsymbol{e}^{\mathrm{policy}}
=
\mathrm{MLP}^{\mathrm{policy}}
\left(
\mathbf{z}_{\pi_{\bar{\theta}}}
\right),
\label{eq:policy_feature}
\end{equation}
where $\mathrm{MLP}^{\mathrm{policy}}(\cdot)$ denotes the policy projection network that maps the hidden policy representation into the extracted policy feature $\boldsymbol{e}^{\mathrm{policy}}$. The fixed policy parameters $\bar{\theta}$ prevent gradients from propagating into the driving policy during scene perturbation network optimization. This design allows the scene perturbation network to leverage the policy representation as auxiliary information while keeping policy optimization decoupled.

These extracted features are then fed into the policy-aware scene encoder:
$(\boldsymbol{e}^{\mathrm{ego}},\boldsymbol{e}^{\mathrm{obj}}_i,\boldsymbol{e}^{\mathrm{road}},\boldsymbol{e}^{\mathrm{risk}},\boldsymbol{e}^{\mathrm{policy}})$.
To model the interaction between the ego vehicle and surrounding objects, TPSP employs an attention module to aggregate object information conditioned on the ego vehicle:
\begin{equation}
\boldsymbol{c}^{\mathrm{obj}}
=
\mathrm{Attention}
\left(
\boldsymbol{e}^{\mathrm{ego}},
\{\boldsymbol{e}_i^{\mathrm{obj}}\}_{i=1}^{N}
\right),
\label{eq:object_attention}
\end{equation}
with $\mathbf{c}^{\mathrm{obj}}$ denoting the aggregated object interaction features and invalid objects masked before attention normalization. The ego feature serves as the query, while surrounding object features provide the interaction information to be aggregated.

Finally, the complete policy-aware scene representation is obtained by fusing all feature components:
\begin{equation}
\boldsymbol{h}
=
\mathrm{MLP}^{\mathrm{fusion}}
\left(
[
\boldsymbol{e}^{\mathrm{ego}},
\boldsymbol{c}^{\mathrm{obj}},
\boldsymbol{e}^{\mathrm{road}},
\boldsymbol{e}^{\mathrm{risk}},
\boldsymbol{e}^{\mathrm{policy}}
]
\right),
\label{eq:scene_policy_representation}
\end{equation}
where $\mathrm{MLP}^{\mathrm{fusion}}(\cdot)$ represents the feature fusion network that combines heterogeneous scene and policy feature into the final policy-aware scene representation $\boldsymbol{h}$. Thus, $\boldsymbol{h}$ captures both the physical interaction context of the scene and the characteristics of the current driving policy. This representation is then used by the subsequent scene perturbation network to generate policy-aware scene perturbations.

\subsection{Targeted Scene Perturbation}

Given the policy-aware scene representation $\boldsymbol{h}$, TPSP generates targeted scene perturbations by first selecting critical objects and then modifying the corresponding scene elements. Instead of applying uniform perturbations to all objects, TPSP focuses on objects that are more relevant to the current policy behavior.

To identify important perturbation targets, TPSP assigns each observable object an importance score. This is achieved by applying a Softmax function over all objects in the scene, which normalizes the raw scoring logits into a probability distribution over the object set. Formally, the importance weight for each object is computed as:
\begin{equation}
\boldsymbol{w}^{\mathrm{obj}}
=
\mathrm{Softmax}
\left(
\left\{
f^{\mathrm{score}}
(
[
\boldsymbol{e}_i^{\mathrm{obj}},
\boldsymbol{h}
]
)
\right\}_{i=1}^{N}
\right),
\end{equation}
where $\boldsymbol{w}^{\mathrm{obj}}$ denotes the normalized importance scores of observable objects, $N$ is the number of objects in the scene, and $f^{\mathrm{score}}(\cdot)$ is an object scoring network that maps each object-scene representation into an importance logit. Based on these scores, TPSP selects the top-$K$ objects as perturbation targets, whose representations are denoted as:
\begin{equation}
\boldsymbol{E}^{\mathrm{obj}}_K
=
\{
\boldsymbol{e}^{\mathrm{obj}}_{1},
\boldsymbol{e}^{\mathrm{obj}}_{2},
\ldots,
\boldsymbol{e}^{\mathrm{obj}}_{K}
\}.
\end{equation}

The selected top-$K$ object representations are then combined with the policy-aware scene representation $\boldsymbol{h}$ and provided to the scene perturbation network $G_{\phi}$. Instead of directly generating deterministic modifications, $G_{\phi}$ models a Gaussian perturbation distribution and samples raw perturbation variables from this distribution.

Specifically, the perturbation network predicts the mean and standard deviation of the Gaussian distribution:
\begin{gather}
\boldsymbol{\mu}
=
g_{\mu}
(
[
\boldsymbol{E}^{\mathrm{obj}}_K,
\boldsymbol{h}
]
),
\\
\boldsymbol{\sigma}
=
\exp
\left(
g_{\sigma}
(
[
\boldsymbol{E}^{\mathrm{obj}}_K,
\boldsymbol{h}
]
)
\right),
\\
\boldsymbol{z}^{\mathrm{raw}}
\sim
\mathcal{N}
\left(
\boldsymbol{\mu},
\mathrm{diag}
(
\boldsymbol{\sigma}^{2}
)
\right),
\label{eq:gaussian_perturbation}
\end{gather}
where $g_{\mu}(\cdot)$ and $g_{\sigma}(\cdot)$ denote the mean and standard deviation prediction heads of $G_{\phi}$, respectively, with exponential function ensuring all standard deviation values positive. $\boldsymbol{\mu}$ and $\boldsymbol{\sigma}$ represent the predicted mean and standard deviation of the perturbation distribution, and $\boldsymbol{z}^{\mathrm{raw}}$ denotes the sampled unconstrained perturbation variables. The diagonal covariance matrix $\mathrm{diag}(\boldsymbol{\sigma}^{2})$ assumes independent sampling across perturbation dimensions.

TPSP then converts the raw perturbation variables into boundary indicators of the final state perturbation,
\begin{equation}
\boldsymbol{z}
=
\lambda
\left(
\boldsymbol{z}^{\mathrm{max}}
\odot
\tanh
(
\boldsymbol{z}^{\mathrm{raw}}
)
\right),
\label{eq:bounded_perturbation}
\end{equation}
where $\boldsymbol{z}^{\mathrm{max}}$ specifies the maximum magnitude of each perturbation variable, and $\odot$ denotes element-wise multiplication. $\lambda$ is a dynamically adjusted scaling factor controlling the overall perturbation strength during training. This transformation converts unconstrained Gaussian samples into physically feasible perturbations while preserving differentiability.

The final perturbation variables $\boldsymbol{z}$ correspond to three types of scene modifications in TPSP: object speed adjustment, object spawn time shift, and ego initial state modification. Object-related perturbations are generated mainly based on the selected object representations, while ego initial state modification is conditioned on the policy-aware scene representation $\boldsymbol{h}$, which captures the current policy's response to the scene, including predicted trajectory distribution or action-value estimates. By integrating object-level and ego-level adjustments in this manner, TPSP generates targeted perturbations rather than fixed or uniform modifications.

\subsection{Threat-guided Scene Perturbation Optimization}

% At the beginning of each training iteration, TPSP aims to improve policy robustness by enriching the training scene set with informative safety-critical experiences. Since the original scene dataset $\mathcal{D}$  may contain limited high-risk interactions, TPSP maintains a scene buffer $\mathcal{B}$ to preserve previously discovered high-threat perturbed scenes and continuously incorporates them into policy training. The two sub-datasets are integrated to form the training dataset $\mathcal{D}_n^{\mathrm{train}}$ for the policy model. To accommodate the requirements of model convergence for scenario distribution across different training stages, the sampling proportions from these datasets are dynamically adjusted by assigning varying weights. 

At the beginning of each training iteration, TPSP enriches the training
scene set with informative safety-critical experiences to improve
policy robustness. Since the original scene dataset $\mathcal{D}$ may
contain insufficient high-risk interactions, TPSP maintains a scene
buffer $\mathcal{B}$ to store previously discovered high-threat
perturbed scenes. The original and perturbed scenes are combined to
construct the training dataset $\mathcal{D}^{\mathrm{train}}$, where
their sampling proportions are dynamically adjusted throughout training.
Specifically, the proportion of high-threat perturbed scenes is
gradually increased during training, allowing the policy to progressively
adapt to more challenging safety-critical situations.

Given any scene sampled from $\mathcal{D}$ or $\mathcal{B}$, the policy model performs rollouts based on the current state. Each rollout is subsequently evaluated from multiple perspectives regarding environmental traffic threats. To evaluate the threat score of a rollout $\tau$, TPSP first computes a step-level threat score:
\begin{equation}
\begin{aligned}
c_t={}&
w^{\mathrm{ttc}}c_t^{\mathrm{ttc}}
+w^{\mathrm{dist}}c_t^{\mathrm{dist}}
+w^{\mathrm{edge}}c_t^{\mathrm{edge}}\\
&+
w^{\mathrm{col}}c_t^{\mathrm{col}}
+w^{\mathrm{off}}c_t^{\mathrm{off}},
\end{aligned}
\label{eq:step_threat}
\end{equation}
where $c_t\in[0,1]$ denotes the threat score at timestep $t$. The five components represent different safety-related factors, including time-to-collision, inter-object distance, road-edge proximity, collision occurrence, and off-road status. The coefficients $w^{\mathrm{ttc}}$, $w^{\mathrm{dist}}$, $w^{\mathrm{edge}}$, $w^{\mathrm{col}}$, and $w^{\mathrm{off}}$ are weighting factors that balance different threat components.

\begin{algorithm}[t]
\caption{TPSP}
\label{alg:past}
\begin{algorithmic}[1]

\STATE Initialize driving policy $\pi_\theta$, scene perturbation network
$G_\phi$, original dataset $\mathcal{D}$, scene buffer $\mathcal{B}$ and perturbation scale $\lambda$

\STATE Set rollout policy
$\pi_{\bar{\theta}}\leftarrow\pi_\theta$

\FOR{each on-policy training iteration $n$}

    \STATE Sample scenes from the combination of the original scene dataset $\mathcal{D}$ and the perturbed scene buffer $\mathcal{B}$.

    \STATE Extract detached policy feature from the sampled scenes using
     the rollout policy $\pi_{\bar{\theta}}$

    \STATE Construct policy-aware scene representations and generate
           targeted scene perturbations with  $G_\phi$

    \STATE Simulate perturbed scenes with
    $\pi_{\bar{\theta}}$ and collect on-policy rollouts

    \STATE Update $\pi_\theta$ using the collected perturbed-scene
    rollouts with the selected on-policy RL algorithm

    \IF{scene perturbation network update}

        \STATE Simulate the same sampled original scenes without perturbations
        using identical simulator seeds

        \STATE Compute threat differences using
        Eq.~\eqref{eq:threat_difference}

        \STATE Update $G_\phi$ by minimizing
        $\mathcal{L}^{\mathrm{per}}$ in Eq.~\eqref{eq:perturb_loss}

    \ENDIF

    \STATE Update $\mathcal{B}$ with high-threat perturbed scenes

    \STATE Adjust $\lambda$ according to the perturbation schedule

    \STATE Synchronize the rollout policy:
    $\pi_{\bar{\theta}}\leftarrow\pi_\theta$

\ENDFOR

\end{algorithmic}
\end{algorithm}

Since safety risks are usually concentrated in a small number of critical moments during a rollout, TPSP summarizes the overall rollout criticality by aggregating the most threatening timesteps. Given a rollout trajectory $\tau=(s_0,a_0,r_0,\ldots,s_T)$ with  rollout horizon $T$, we select the top-$k$ timesteps with the highest step-level threat scores and compute the rollout-level criticality as:
\begin{equation}
\mathcal{C}(\tau)
=
\frac{1}{k'}
\sum_{t\in\mathrm{Top}(\tau, k)}
c_t,
\quad
k'=\min(k,T+1),
\label{eq:rollout_criticality}
\end{equation}
where $\mathrm{Top}(\tau, k)$ retrieves the temporal indices of the top-$k$ selected critical timesteps with the highest $c_t$ values. Since a rollout of horizon $T$ contains $T+1$ states, $k'$ adjusts the number of selected timesteps when the rollout length is shorter than $k$. By focusing on the most threatening moments rather than averaging all timesteps, $\mathcal{C}(\tau)$ better captures the safety-critical characteristics of a rollout.

In addition to the overall threat magnitude, TPSP considers the temporal instability of threat evolution during the rollout. While $\mathcal{C}(\tau)$ captures critical risk levels, it does not reflect how rapidly the threat changes over time. Therefore, TPSP introduces the threat instability metric:
\begin{equation}
\mathcal{V}(\tau)
=
\frac{1}{T}
\sum_{t=0}^{T-1}
|c_{t+1}-c_t|,
\label{eq:threat_variation}
\end{equation}
where $\mathcal{V}(\tau)$ measures the temporal instability of the threat evolution over the rollout. A larger value indicates that the safety condition changes more rapidly, suggesting a more challenging interaction for the driving policy.

Based on both threat severity and threat instability, the final threat score of a rollout is defined as:
\begin{equation}
\mathcal{\xi}(\tau)
=
\mathcal{C}(\tau)
\frac{1+\beta\mathcal{V}(\tau)}
{1+\beta},
\label{eq:threat_level}
\end{equation}
where $\beta$ controls the contribution of threat variation. 
The criticality term $\mathcal{C}(\tau)$ dominates the threat estimation to prioritize genuinely safety-critical scenes, while $\mathcal{V}(\tau)$ acts as a modulation factor to emphasize temporally unstable interactions. This formulation enables TPSP to identify informative safety-critical scenes for policy improvement.

\begin{figure}[t]
    \centering
    \includegraphics[width=0.46\textwidth]{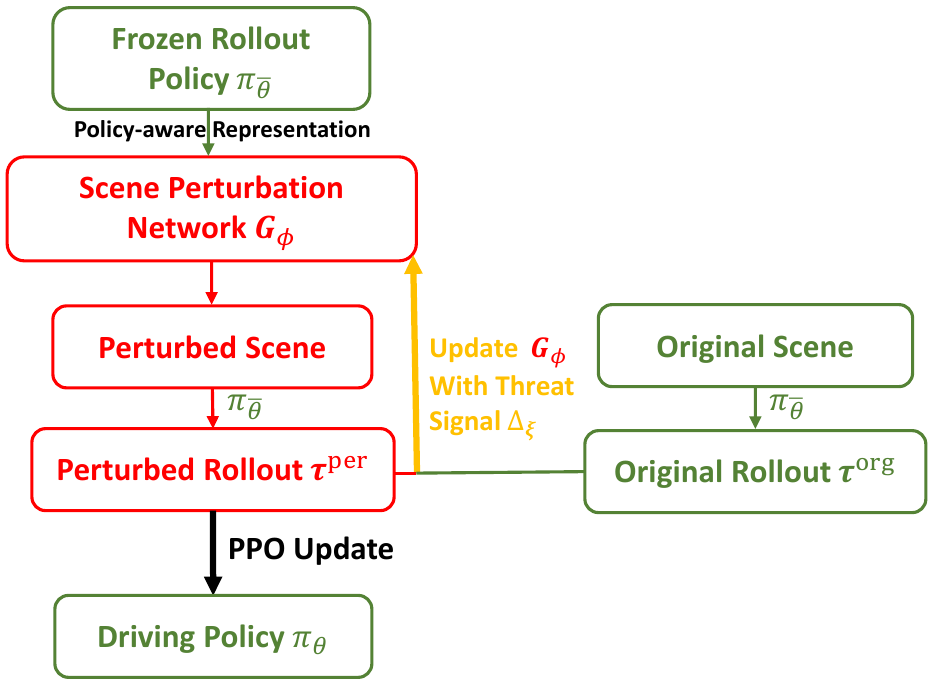}
    \caption{Closed-loop optimization between the driving policy and scene perturbation network in TPSP.}
    \label{fig:optimization}
    \vspace{-4mm}
\end{figure}

To optimize the scene perturbation network, TPSP compares the threat scores  between policy rollouts on perturbed and original scenes from the same initial current state. Specifically, both rollouts are generated using the same fixed driving policy $\pi_{\bar{\theta}}$ and identical simulator seeds:
\begin{equation}
\Delta\mathcal{\xi}
=
\mathcal{\xi}
(
\tau^{\mathrm{per}}
)
-
\mathcal{\xi}
(
\tau^{\mathrm{org}}
),
\label{eq:threat_difference}
\end{equation}
where $\tau^{\mathrm{per}}$ and $\tau^{\mathrm{org}}$ denote the perturbed and original rollouts of the same initial state, respectively. The threat difference $\Delta\mathcal{\xi}$ measures the additional safety difficulty introduced by the generated perturbation.

The perturbation network $G_{\phi}$ is optimized using the threat difference as the learning signal. Based on one certain initial scene or state $s\in\mathcal{S}$, for a sampled perturbation vector $\boldsymbol z_s^{\mathrm{raw}}$, its log probability is defined as:
\begin{equation}
\log p_{\phi}
(
\boldsymbol z_s^{\mathrm{raw}}
|
\boldsymbol{E}^{\mathrm{obj}}_K,
\boldsymbol{h}
).
\label{eq:perturb_logprob}
\end{equation}

\begin{figure*}[t]
    \hspace{-6mm}
    \centering
    \begin{subfigure}{0.5\textwidth}
        \centering
        \includegraphics[width=\linewidth]{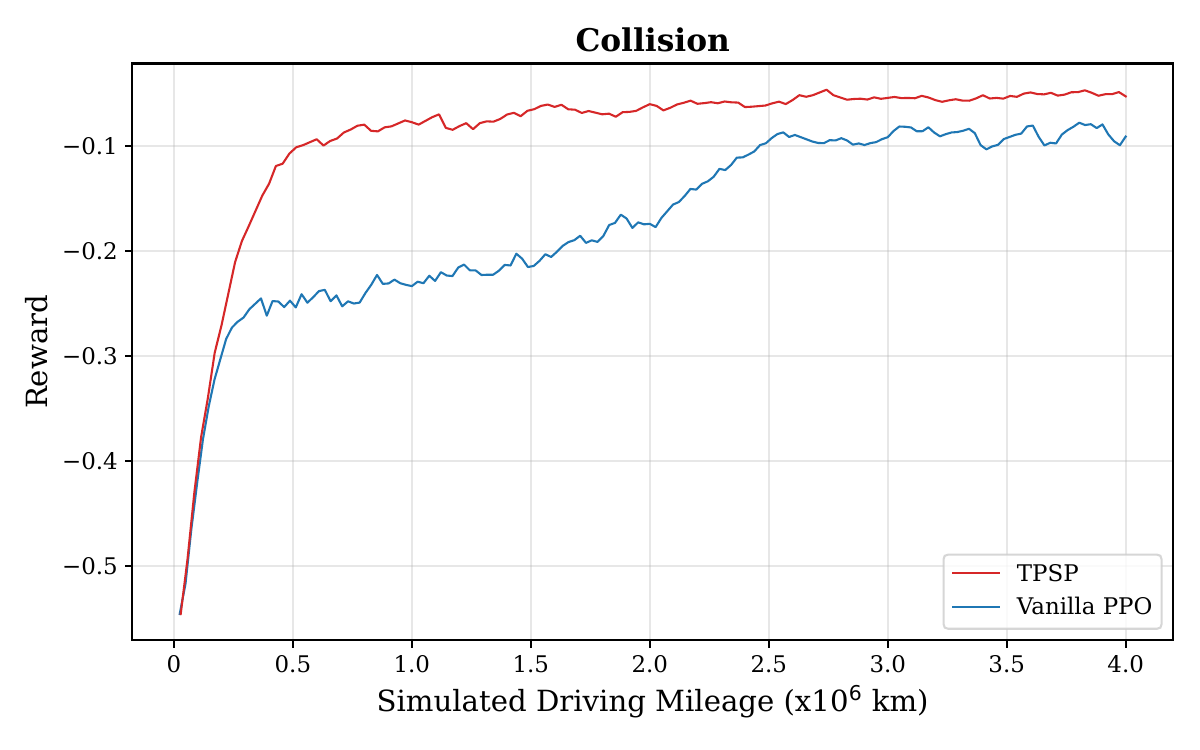}
        \label{fig:collision_reward}
    \end{subfigure}
    \hfill
    \begin{subfigure}{0.5\textwidth}
        \centering
        \includegraphics[width=\linewidth]{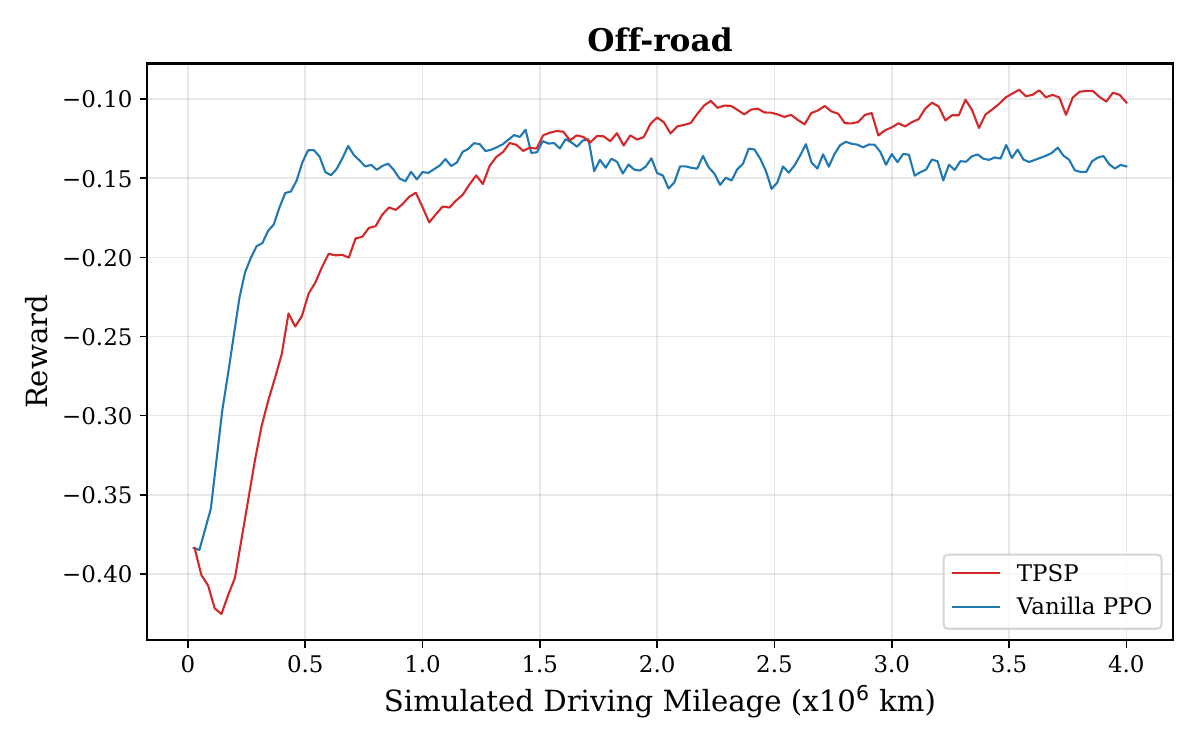}
        \label{fig:offroad_reward}
    \end{subfigure}
    \vspace{-7mm}
    \caption{
    Safety learning reward curves during online RL training. Here, collision reward penalizes vehicle collision events,
while off-road reward penalizes curb collisions. TPSP achieves
    faster improvements in collision avoidance and off-road safety with fewer simulated driving mileage.
    }
    \label{fig:safety_learning_curve}
\end{figure*}

\begin{figure*}[t]
    \centering
    \vspace{-2mm}
    \setlength{\tabcolsep}{2pt}
    \begin{tabular}{@{}cccc@{}}
        \multicolumn{2}{c}{\textbf{Scene A}} &
        \multicolumn{2}{c}{\textbf{Scene B}} \\[0.5mm]
        \includegraphics[width=0.235\textwidth,trim=6 6 6 6,clip]{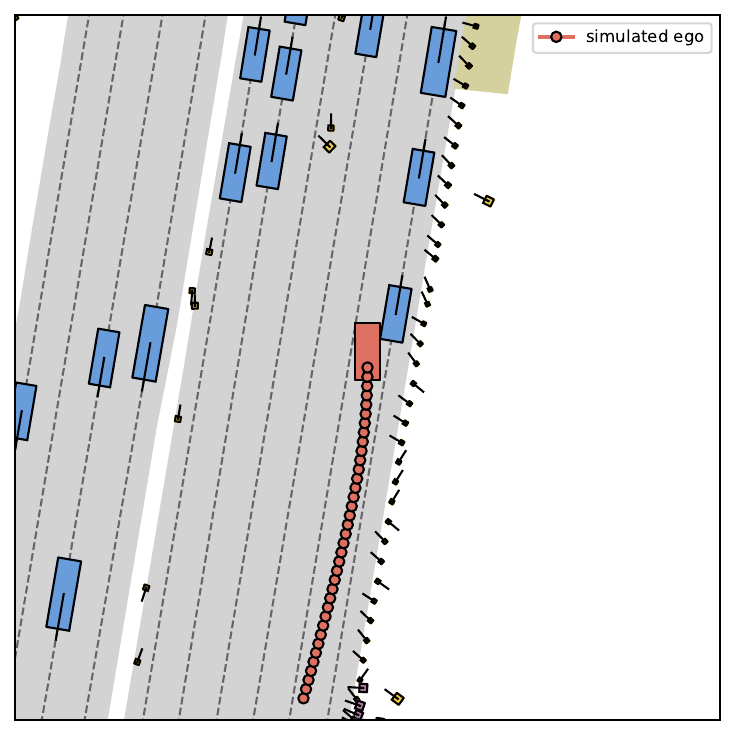}
        &
        \includegraphics[width=0.235\textwidth,trim=6 6 6 6,clip]{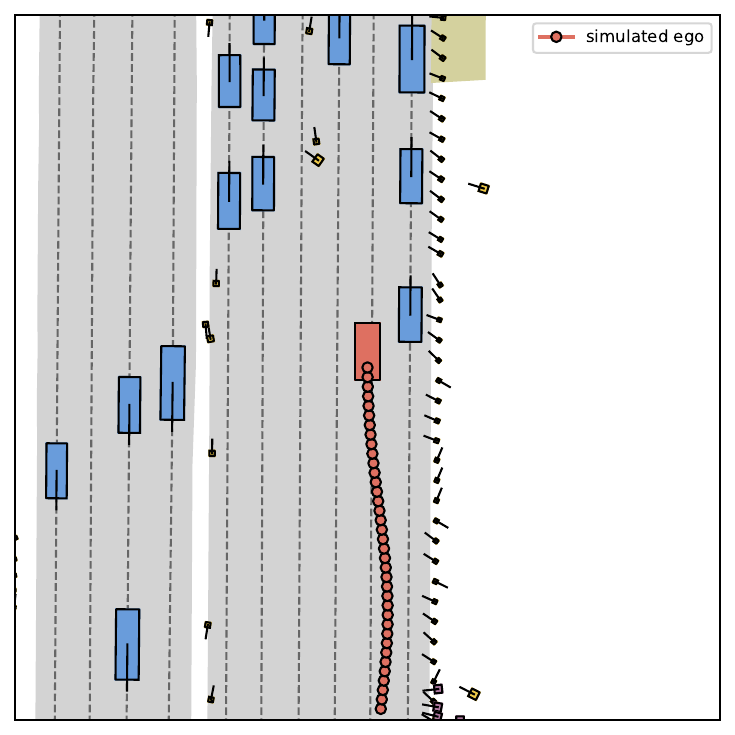}
        &
        \includegraphics[width=0.235\textwidth,trim=6 6 6 6,clip]{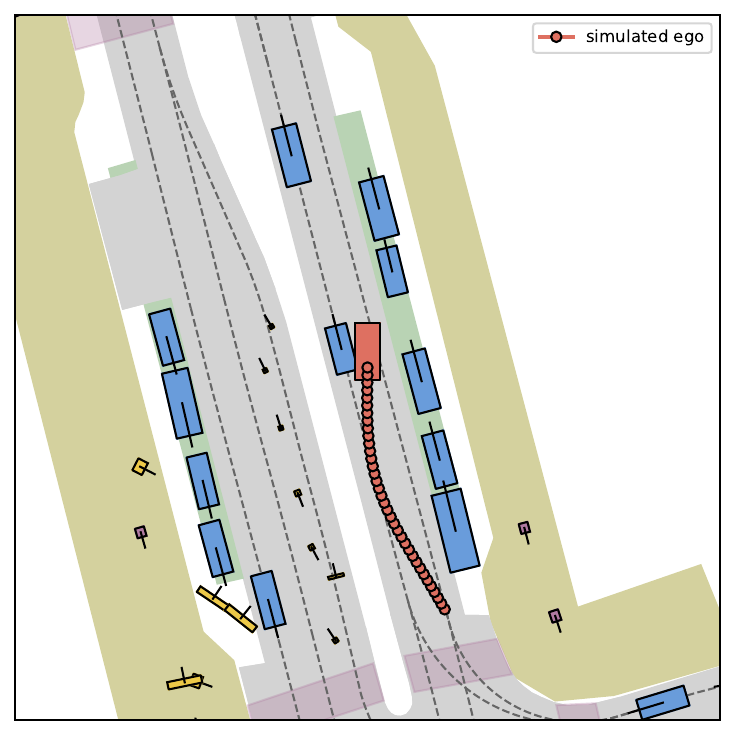}
        &
        \includegraphics[width=0.235\textwidth,trim=6 6 6 6,clip]{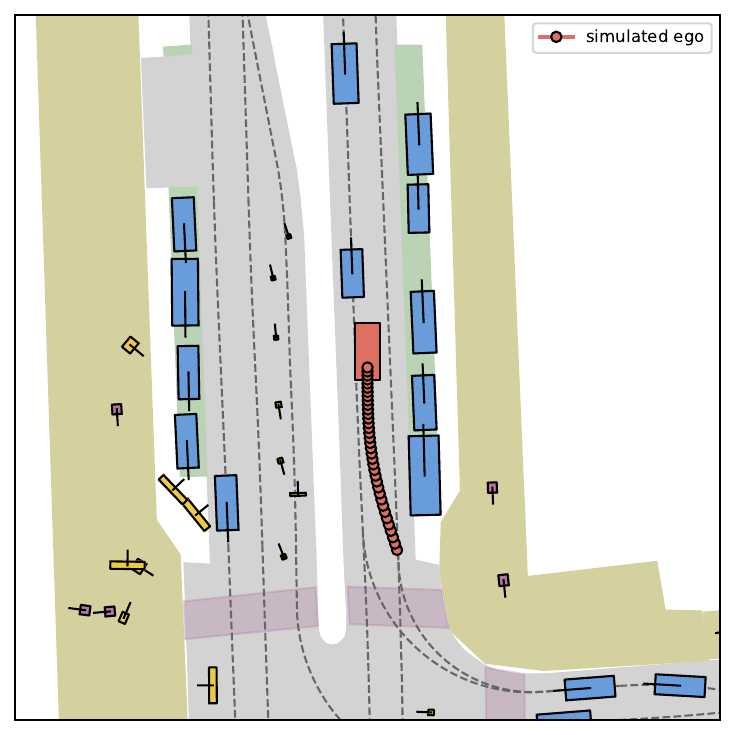}
        \\[-1mm]
        \small Vanilla PPO &
        \small TPSP &
        \small Vanilla PPO &
        \small TPSP
    \end{tabular}
    \vspace{-2mm}
    \caption{Qualitative comparison on two representative safety-critical NAVSIM v2 scenes. TPSP enables earlier hazard response and safer trajectory generation compared with Vanilla PPO.}
    \label{fig:qualitative_navsim}
    \vspace{-1mm}
\end{figure*}

The threat differences are normalized within each perturbation update batch to reduce the scale variation across training iterations. The normalized threat advantage for scene $s$ is computed as:
\begin{equation}
\hat{A}_{s}
=
\frac{
\Delta \xi_{s}
-
\operatorname{mean}
(
\Delta \xi
)
}
{
\operatorname{std}
(
\Delta \xi
)
+\epsilon
},
\label{eq:threat_advantage}
\end{equation}
where $\hat{A}_{s}$ denotes the normalized threat advantage of scene $s$. The operators $\operatorname{mean}(\cdot)$ and $\operatorname{std}(\cdot)$ compute the mean and standard deviation of the threat differences over the current perturbation update batch, respectively, and $\epsilon$ is a small constant for numerical stability. Thus, the scene perturbation network is optimized with:
\begin{equation}
\mathcal L^{\mathrm{per}}
=
-
\mathbb E_{s}
[
\hat A_{s}
\log p_{\phi}
(
\boldsymbol z^{\mathrm{raw}}_{s}
|
\boldsymbol{E}^{\mathrm{obj}}_K,
\boldsymbol{h}
)
]
-
c^{\mathrm{ent}}
H_{\boldsymbol{z}},
\label{eq:perturb_loss}
\end{equation}
where $c^{\mathrm{ent}}$ is the entropy regularization coefficient and $H_{\boldsymbol{z}}$ denotes the entropy of the perturbation distribution. The entropy regularization encourages exploration of the perturbation space and prevents premature convergence.

Algorithm~\ref{alg:past} and Figure~\ref{fig:optimization} illustrate the overall closed-loop optimization procedure of TPSP. During training, the generated perturbed scenes are used to collect rollouts for updating the driving policy $\pi_\theta$. Meanwhile, the threat difference between perturbed and original rollouts provides the optimization signal for updating the scene perturbation network $G_\phi$. By iteratively improving the driving policy based on informative perturbed scenes with PPO~\citep{ schulman2017ppo}  algorithm and refining  $G_\phi$ with threat-guided signals, TPSP establishes a closed-loop optimization process for discovering safety-critical training scenes.

\section{Experiments}
\label{sec:experiments}

Our experiments assess TPSP’s safety and efficiency from three perspectives: learning speed, perturbation efficacy, and final performance. We first determine if TPSP improves safety behaviors faster under fixed interaction budgets. Next, ablation studies isolate the impact of targeted perturbations and policy-aware representations. We then compare TPSP’s final policy against baselines on the NAVSIM v2 benchmark. Qualitative results further illustrate that TPSP effectively optimizes scene
perturbations using threat signals from policy rollouts to generate
informative safety-critical interactions.

\subsection{Experimental Setup}
We train TPSP using the GPUDrive simulator with PPO and the NAVSIM v2 \texttt{navtrain} dataset. The resulting driving policies
are evaluated on the challenging NAVSIM v2
\texttt{navhard\_two\_stage} benchmark. The evaluation considers three safety-related metrics: no at-fault collisions (NC), drivable area
compliance (DAC), and time to collision (TTC). During training, all
methods share the same policy architecture, reward function, interaction
budgets, and PPO optimization hyperparameters to ensure a fair
comparison. Further implementation details are provided in Appendix A.

\subsection{Safety Learning Efficiency Analysis}
Figure~\ref{fig:safety_learning_curve} illustrates the evolution of
safety-related rewards during online RL training under the same
interaction budgets. TPSP achieves faster improvements in both collision
avoidance and off-road safety rewards compared with Vanilla PPO,
indicating that the policy can acquire safety-related behaviors more
efficiently with fewer simulated driving miles. For example, at 2M miles, TPSP already reaches a collision-reward level that Vanilla PPO does not attain until 4M miles. This improvement comes
from the targeted exposure to informative safety-critical interactions,
which provides more effective training signals than uniformly sampled
experiences.

To further analyze the learned safety behaviors, Figure~\ref{fig:qualitative_navsim} presents two safety-critical scenarios from NAVSIM v2 to analyze learned safety behaviors. Under identical initial conditions, TPSP produces safer trajectories with earlier hazard responses. In Scene A, TPSP proactively avoids a collision by adjusting its trajectory, while Vanilla PPO fails to react and crashes. In Scene B, TPSP handles abrupt braking by decelerating and keeping a safe distance, whereas Vanilla PPO responds too late. These cases highlight TPSP’s ability to improve safety learning efficiency by focusing on challenging interactions relevant to the policy’s weaknesses.

\begin{table}[H]
\centering
\caption{
Ablation on NAVSIM v2
\texttt{navhard\_two\_stage}.
}
\label{tab:ablation}
\vspace{-2mm}
\footnotesize
\setlength{\tabcolsep}{3.8pt}
\renewcommand{\arraystretch}{1.05}
\resizebox{1.0\columnwidth}{!}{
\begin{tabular}{lcccccc}
\toprule
\multirow{2}{*}{Method}
& \multicolumn{2}{c}{NC $\uparrow$}
& \multicolumn{2}{c}{DAC $\uparrow$}
& \multicolumn{2}{c}{TTC $\uparrow$} \\
\cmidrule(lr){2-3}
\cmidrule(lr){4-5}
\cmidrule(lr){6-7}
& S1 & S2 & S1 & S2 & S1 & S2 \\
\midrule

Vanilla PPO
& 96.2 & 85.7
& 92.4 & 81.0
& 96.0 & 82.4 \\

Random Perturbation
& 97.1 & 83.2
& 94.4 & 83.8
& 96.8 & 80.3 \\

TPSP w/o PA
& 98.8 & 91.1
& 97.5 & 90.4
& 98.8 & 89.7 \\

\textbf{TPSP}
& \textbf{99.8} & \textbf{96.7}
& \textbf{97.8} & \textbf{93.7}
& \textbf{99.6} & \textbf{94.6} \\
\bottomrule
\vspace{-4mm}
\end{tabular}
}
\end{table}

\subsection{Ablation Study}

We conduct ablation studies to analyze the contribution of different
components in TPSP. We compare Vanilla PPO, random perturbation,
TPSP without policy awareness (TPSP w/o PA), and the full TPSP framework with
policy-aware targeted perturbation. Table~\ref{tab:ablation} reports the safety performance on the NAVSIM v2
\texttt{navhard\_two\_stage} benchmark. Here, S1 and S2 denote Stage 1
and Stage 2 evaluations, respectively, where Stage 1 evaluates policies
on original scenes and Stage 2 focuses on more challenging synthesized
scenarios. The upward arrows indicate that higher values correspond to
better performance for the corresponding safety metrics.

TPSP achieves the best performance across both stages, particularly in
the more challenging S2 setting, where it obtains 96.7\% NC and 94.6\%
TTC. Random perturbation provides only limited improvements over
Vanilla PPO, indicating that increasing scene diversity alone is
insufficient for effective safety learning. Moreover, the performance
gap between TPSP and TPSP w/o PA demonstrates that
incorporating policy-specific information is essential for generating
more informative safety-critical scenes.

\subsection{Safety Evaluation on NAVSIM v2}

We show the safety of TPSP by comparing it with state-of-the-art methods on the NAVSIM v2 leaderboard (Table~\ref{tab:navsim_safety}). TPSP achieves superior performance in NC and TTC metrics across both evaluation stages, highlighting its effectiveness in collision avoidance and interaction safety. Further details on the baseline methods are available in Appendix B.

\begin{table}[H]
\centering
\caption{
 Safety on NAVSIM v2
\texttt{navhard\_two\_stage}.
External results are reported from the NAVSIM v2 benchmark leaderboard for reference.~\citep{cao2025navsimv2}
}
\label{tab:navsim_safety}
\vspace{-2mm}
\footnotesize
\setlength{\tabcolsep}{3.8pt}
\renewcommand{\arraystretch}{1.05}
\resizebox{0.85\columnwidth}{!}{
\begin{tabular}{lcccccc}
\toprule
\multirow{2}{*}{Method}
& \multicolumn{2}{c}{NC $\uparrow$}
& \multicolumn{2}{c}{DAC $\uparrow$}
& \multicolumn{2}{c}{TTC $\uparrow$} \\
\cmidrule(lr){2-3}
\cmidrule(lr){4-5}
\cmidrule(lr){6-7}
& S1 & S2 & S1 & S2 & S1 & S2 \\
\midrule

NavFormer
& 96.2 & 85.7
& 92.4 & 81.0
& 96.0 & 82.4 \\

RAP
& 97.1 & 83.2
& 94.4 & 83.8
& 96.8 & 80.3 \\

ZTRS
& 98.8 & 91.1
& 97.5 & 90.4
& 98.8 & 89.7 \\

SimScale
& 99.5 & 94.5
& \textbf{99.1} & \textbf{94.2}
& 99.5 & 92.8 \\

DrivoR
& 99.1 & 92.3
& 98.2 & 91.6
& 98.6 & 90.5 \\
\midrule
\textbf{TPSP}
& \textbf{99.8} & \textbf{96.7}
& 97.8 & 93.7
& \textbf{99.6} & \textbf{94.6} \\
\bottomrule
\vspace{-6mm}
\end{tabular}
}
\end{table}

\begin{figure}[H]
    \centering
    \includegraphics[width=0.42\columnwidth, height=7.8cm, keepaspectratio=false]{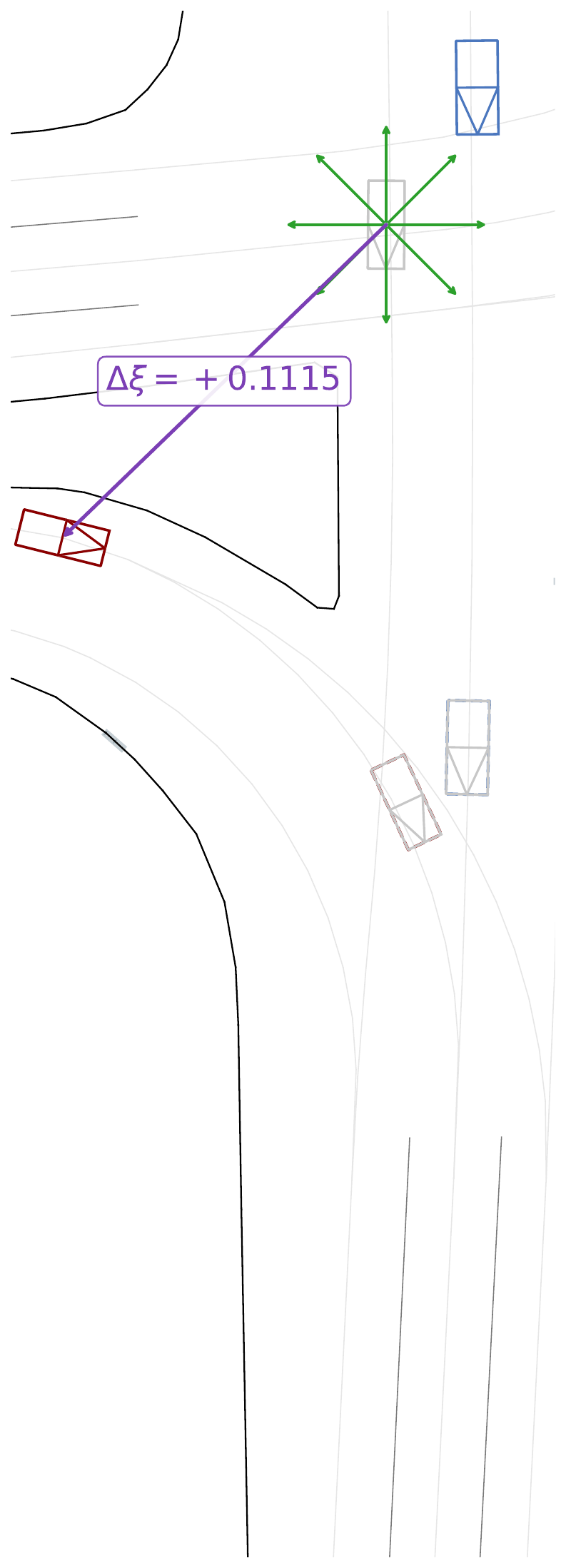}
    \hspace{0.02\columnwidth}
    \includegraphics[width=0.42\columnwidth, height=7.8cm, keepaspectratio=false]{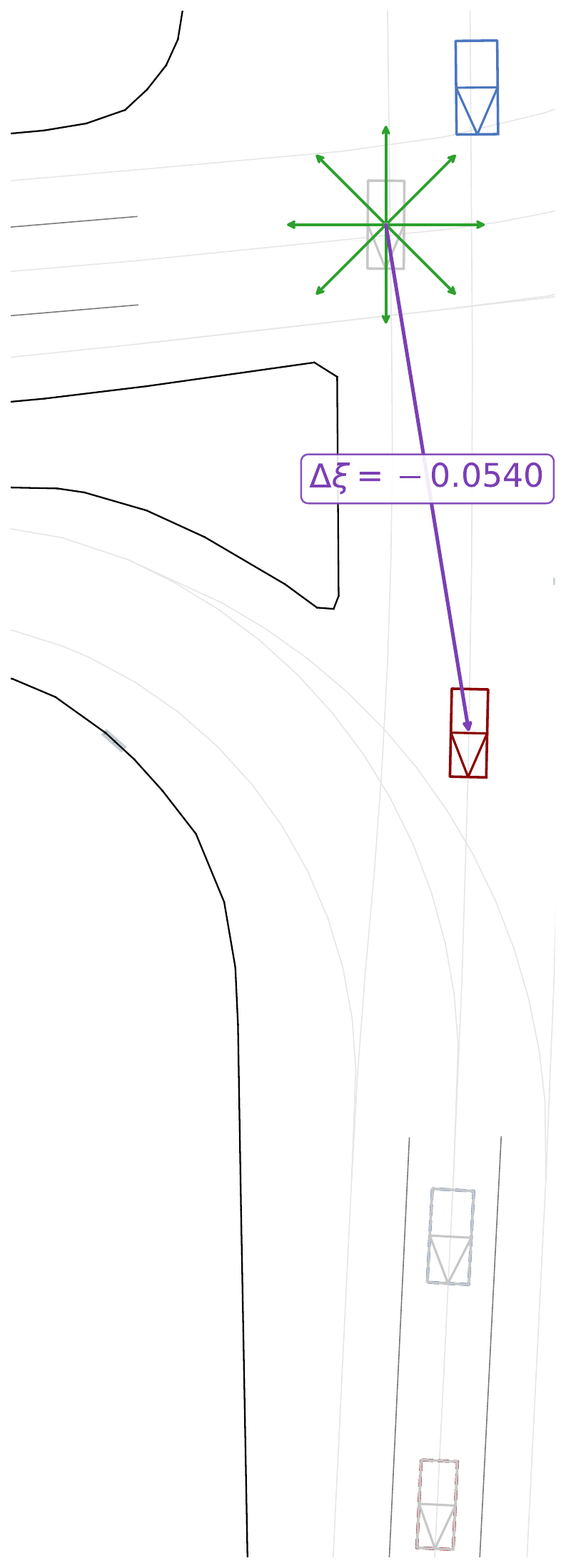}
    \vspace{-2mm}
    \caption{Visualization of threat-guided policy-aware perturbations learned by TPSP in GPUDrive simulation. Left: a cut-in situation; Right: a car-following situation. TPSP evaluates how different perturbations affect policy-rollout threat signals and optimizes scene perturbations toward safety-critical interactions.}
    \label{fig:perturbation_vis}
    \vspace{-2mm}
\end{figure}

\subsection{Qualitative Analysis}

Figure~\ref{fig:perturbation_vis} visualizes the learned perturbations generated by TPSP in GPUDrive simulation. In the cut-in situation (left), the generated perturbation increases the threat of the interaction with $\Delta\xi=+0.1115$, corresponding to a more dangerous merging conflict. In contrast, the car-following situation (right) exhibits a less hazardous interaction pattern, and the estimated threat variation $\Delta\xi=-0.0540$ is consistent with this intuitive observation. These examples demonstrate that TPSP can effectively evaluate the risk variation induced by different perturbations and optimize scene modifications toward more informative safety-critical interactions, rather than blindly increasing scene complexity.

\section{Conclusion and Future Work}

We proposed TPSP, a Threat-guided Policy-aware Scene Perturbation framework for improving safety learning efficiency in online RL for autonomous driving. By generating targeted perturbations guided by policy weaknesses, TPSP enables more informative exploration under limited training budgets. Experiments on NAVSIM v2 demonstrate that TPSP achieves strong safety performance with approximately 4 million kilometers of simulated driving mileage. Ablation studies further validate the effectiveness of policy-aware targeted perturbations. Future work will explore more comprehensive perturbation spaces beyond the current scene-level modifications, including richer semantic, map-level, and multi-agent interaction perturbations. We also plan to investigate TPSP in larger-scale simulation environments and real-world driving settings, as well as its integration with more diverse RL
algorithms and autonomous driving policy architectures.

\bibliography{aaai2027}

\clearpage
\appendix

\section{Appendix A. Implementation Details}
\label{sec:implementation_details}

This section provides additional implementation details of TPSP, including
training hyperparameters, training infrastructure and
safety-related reward implementation. These details are provided to facilitate
the reproducibility of our experiments.

\subsection{Training Hyperparameters}
\label{sec:hyperparameters}

TPSP consists of a driving policy $\pi_{\theta}$ optimized with PPO and a
scene perturbation network $G_{\phi}$ trained through threat-guided
optimization. The main configurations of these components are summarized
below.

\subsubsection{Driving Policy Configuration}

The driving policy $\pi_{\theta}$ is optimized using PPO in the GPUDrive
environment. The main architecture and training configurations are summarized
in Table~\ref{tab:policy_architecture}.

\begin{table}[H]
\centering
\caption{Architecture configuration of the driving policy.}
\label{tab:policy_architecture}
\small
\begin{tabular}{lc}
\toprule
Hyperparameter & Value \\
\midrule
Network architecture & MLP-based actor network \\
Hidden dimension & 64 \\
Latent policy dimension & 64 \\
Activation function & $\tanh$ \\
Action space & Discrete \\
Steering bins & 13 \\
Acceleration bins & 7 \\
\bottomrule
\end{tabular}
\end{table}

\subsubsection{Scene Perturbation Network Configuration}

The scene perturbation network $G_{\phi}$ generates policy-aware scene
modifications based on encoded simulator states and detached policy features.
The main configurations are summarized in Table~\ref{tab:perturb_hyperparameters}.

\begin{table}[H]
\centering
\caption{Configuration of the scene perturbation network.}
\label{tab:perturb_hyperparameters}
\small
\begin{tabular}{lc}
\toprule
Hyperparameter & Value \\
\midrule
Encoder hidden dimension & 128 \\
Fusion dimension & 256 \\
Policy feature dimension & 64 \\
Number of selected objects $K$  & 8 \\
Initial log standard deviation & -1.0 \\
Optimizer & Adam \\
Learning rate & $1\times10^{-4}$ \\
Entropy coefficient & 0.01 \\
Gradient clipping & 0.5 \\
Update interval & 4 rollouts \\
Initial perturbation scale $\lambda$ & 0.2 \\
\bottomrule
\end{tabular}
\end{table}

\subsubsection{PPO Optimization Configuration}

The detailed PPO optimization parameters used for updating the driving policy
are summarized in Table~\ref{tab:ppo_hyperparameters}.

\begin{table}[H]
\centering
\caption{PPO optimization configuration for training the driving policy.}
\label{tab:ppo_hyperparameters}
\small
\begin{tabular}{lc}
\toprule
Hyperparameter & Value \\
\midrule
Optimizer & AdamW \\
Learning rate & $3\times10^{-4}$ \\
PPO clip ratio & 0.2 \\
Number of PPO epochs & 2 \\
Discount factor $\gamma$ & 0.99 \\
GAE coefficient $\lambda_{\mathrm{GAE}}$ & 0.95 \\
Experience batch size & 204800 \\
Mini-batch size & 6400 \\
Value loss coefficient & 0.5 \\
Entropy coefficient & 0.01 \\
Gradient clipping & 0.5 \\
\bottomrule
\end{tabular}
\end{table}

\subsection{Training Infrastructure}
\label{sec:training_infrastructure}

All experiments are conducted on the GPUDrive simulation platform.
The hardware and software configurations used for training are summarized
in Table~\ref{tab:hardware_configuration}.

\begin{table}[H]
\centering
\caption{Training infrastructure configuration.}
\label{tab:hardware_configuration}
\small
\begin{tabular}{lc}
\toprule
Configuration & Specification \\
\midrule
GPU & NVIDIA A100 \\
Number of GPUs & 16 \\
Memory & 1024GB \\
Number of CPU Cores & 128 \\
Operating System & Linux \\
Deep learning framework & PyTorch \\
Simulation backend & GPUDrive \\
Parallel simulation environments & 512 \\
\bottomrule
\end{tabular}
\end{table}

\subsection{Safety-related Reward Implementation}
\label{sec:reward_details}

The driving policy is optimized using the reward provided by the GPUDrive
simulator. This section describes the implementation of two safety-related
reward components: collision and off-road penalties. At each simulation step, GPUDrive provides safety event indicators for each
agent. The collision signal combines collisions with vehicles and other
objects:
\begin{equation}
d_t^{\mathrm{col}}
=
d_t^{\mathrm{veh}}
+
d_t^{\mathrm{obj}},
\end{equation}
where $d_t^{\mathrm{veh}}$ and $d_t^{\mathrm{obj}}$ denote collision events
with vehicles and other objects at timestep $t$, respectively. The off-road
signal $d_t^{\mathrm{off}}$ indicates whether the ego vehicle leaves the
drivable area. The corresponding safety-related reward terms are defined as:
\begin{equation}
r_t^{\mathrm{safe}}
=
-3.5d_t^{\mathrm{col}}
-0.75d_t^{\mathrm{off}}
+0.5d_t^{\mathrm{goal}},
\end{equation}
where $d_t^{\mathrm{goal}}$ denotes goal achievement. The larger collision
penalty encourages the policy to prioritize collision avoidance, while the
off-road penalty provides continuous guidance for maintaining valid driving
areas.

The collision penalty is applied only at the collision timestep since collided
agents are removed by the simulator. In contrast, off-road penalties can
accumulate when the vehicle remains outside the drivable area. Other shaping
reward components provided by GPUDrive remain unchanged during training.

\section{Appendix
 B. Additional Discussion on NAVSIM Comparison and Training Pipeline}
\label{sec:navsim_appendix}

This section further discusses the comparison between TPSP and existing
NAVSIM~v2 leaderboard methods, and describes the training pipeline used to
enable online reinforcement learning with NAVSIM scenes. We clarify the
differences in learning paradigms and input representations between TPSP and
existing benchmark approaches.

\subsection{Comparison with NAVSIM v2 Leaderboard Methods}

Table~\ref{tab:navsim_safety} compares TPSP with representative methods
reported on the official NAVSIM~v2 leaderboard. These external results are
included for reference under the same benchmark evaluation protocol.

Most existing NAVSIM leaderboard methods follow an offline data-driven
end-to-end autonomous driving paradigm\citep{cao2025navsimv2,feng2026rap,li2025ztrs,Tian2026Sim,kirby2026drivor}. They typically learn driving policies
or trajectory planners from large-scale recorded driving data, where raw
sensor observations provided by NAVSIM, such as multi-view camera inputs,
are directly used as model inputs. Therefore, these approaches generally
adopt a one-stage learning pipeline that maps perception-level observations
to future trajectories or driving actions.

In contrast, TPSP is designed as an online reinforcement learning framework
that improves policy learning through interactive simulation. Instead of
directly consuming raw sensor observations, TPSP utilizes structured
white-box information available from the simulator, including ego states,
surrounding object states, road information, and risk-related features. Based
on these structured representations, TPSP first constructs a policy-aware
scene representation and then performs online RL optimization with targeted
scene perturbations.

Therefore, TPSP differs from existing NAVSIM methods in both learning
paradigm and system pipeline. Existing approaches mainly improve driving
performance by developing stronger perception and planning models from
offline data, whereas TPSP focuses on improving the quality of online policy
training experiences through simulator-based interaction. Despite these
differences, TPSP achieves competitive performance on NAVSIM~v2 safety
evaluation, demonstrating the effectiveness of online RL with policy-aware
scene optimization.

\subsection{NAVSIM-to-GPUDrive Training Pipeline}

TPSP requires interactive simulation for online reinforcement learning,
whereas NAVSIM v2 provides recorded driving scenes for benchmark evaluation.
Therefore, the original NAVSIM v2 scenes cannot be directly used for
GPUDrive simulation and training.

To enable online training, we transform NAVSIM v2 scenes into GPUDrive-compatible
simulation environments. The conversion process preserves essential scene
information, including map structures, dynamic agent states, and temporal
interactions, while enabling large-scale parallel simulation in GPUDrive.

After conversion, TPSP performs online reinforcement learning on the generated
GPUDrive scenes. The learned policy is finally evaluated on the official
NAVSIM~v2 \texttt{navhard\_two\_stage} benchmark following the standard
evaluation protocol, allowing comparison with the publicly reported
leaderboard results.
\end{document}